\documentclass{article}
\usepackage{amsmath}
\PassOptionsToPackage{numbers}{natbib}
\usepackage[utf8]{inputenc}

\usepackage[preprint]{neurips_2025}
\usepackage[T1]{fontenc}

\usepackage{fontenc}    
\usepackage{hyperref}       
\usepackage{url}            
\usepackage{booktabs}       
\usepackage{amsfonts}       
\usepackage{nicefrac}       
\usepackage{microtype}      
\usepackage{xcolor}         
\usepackage{graphicx}       
\usepackage{caption}        
\usepackage{subcaption}     

\title{AbideGym: Turning Static RL Worlds into Adaptive Challenges}

\author{%
Abi Aryan,
Zac Yung-Chun Liu,
Aaron Childress \\
Abide AI
}
\begin{document}

\maketitle

\begin{abstract}
Agents trained with reinforcement learning often develop brittle policies that fail when dynamics shift, a problem amplified by static benchmarks. AbideGym, a dynamic MiniGrid wrapper, introduces agent-aware perturbations and scalable complexity to enforce intra-episode adaptation. By exposing weaknesses in static policies and promoting resilience, AbideGym provides a modular, reproducible evaluation framework for advancing research in curriculum learning, continual learning, and robust generalization.
\end{abstract}

\section{Introduction}
Developing agents that generalize robustly across diverse and changing environments remains a central challenge. While deep RL has achieved remarkable success in complex domains such as games and simulated control, these accomplishments often rely on static, narrowly defined environments, resulting in brittle policies that fail when exposed to even slight variations unseen during training \cite{packer2019assessing}. This generalization gap is a critical barrier to deploying RL systems in the real world, where conditions are rarely stationary.

A growing body of research highlights this deficiency. For instance, \citet{cobbe2019} introduced the CoinRun benchmark, which uses procedural content generation (PCG) to create distinct training and test sets, revealing that standard deep RL agents overfit to surprisingly large training sets. \citet{packer2019assessing} further suggests that increasing environmental diversity during training is a key factor for robust learning, sometimes proving more effective than purpose-built generalization algorithms. More recent studies have reinforced these findings: \citet{generalization_gap_offline2024} and \citet{generalization_multiobjective2025} demonstrate that offline and multi-objective RL agents struggle to transfer policies to even moderately altered environments, highlighting the limitations of static training regimes. Similarly, \citet{action_adaptive2025} and \citet{causal_transfer2025} show that agents frequently fail to adapt when action spaces or environmental rules change, indicating that most benchmarks evaluate generalization only across episodes, not within episodes. This gap is critical because real-world scenarios often involve intra-episode shifts like tools breaking, rules changing, or competitors altering strategies which require agents to revise their policy dynamically and maintain performance under evolving conditions \citep{deep_rl_not_human2025, exploration_generalization2024, procgen_generalization2024}.

To address this limitation, we introduce \textbf{AbideGym}, a dynamic environment wrapper that transforms static Gym \cite{gymnasium} tasks into adaptive, evolving perturbations \cite{aryan2025}. Rather than providing new environments from scratch, AbideGym augments existing environments with agent-aware, structured variability to address the problem of limited testbeds for benchmarking. Our initial implementation extends the classic door-key-goal task \cite{minigrid} with two key mechanisms:

\begin{enumerate}
\item \textbf{Timeout-based perturbations:} The environment tracks agent inactivity. If an agent remains idle or fails to progress, the task rules are modified. The intuition behind this is \textit{state pertubation} \cite{aryan2025}, implying that there has been a change in state like major time-elapsed since the agent activity which results in \textit{hesitation (or confidence) penalty}. For example, when this happens, the key will now no longer unlock the door, and an alternative solution path (a "trigger tile") appears. \textit{This encourages agents to abandon ineffective policies and explore new ones also known in the literature as the classic Exploration versus Exploitation problem.}
\item \textbf{Dynamic resizing:} The environment proactively increases its spatial complexity by expanding the grid and introducing additional obstacles. This allows tasks that agents have already mastered to remain challenging and prevents overfitting to a single configuration, effectively supporting curriculum-based difficulty scaling. \textit{The key intuition behind this is to train the agents for policy adaptation instead of pure policy memorization.}
\end{enumerate}

AbideGym’s core distinction is that environmental changes are not random or pre-scheduled but are triggered by the agent’s behavior. This ensures structured, reproducible, and context-aware variability in the environment. Our contributions are threefold:

\begin{itemize}
\item We introduce a dynamic environment framework that injects controlled, agent-aware, intra-episode variability into standard RL benchmarks, turning static tasks into adaptive challenges.
\item We provide a flexible platform for exploring adaptive behaviors, including strategy switching, generalization under increasing task complexity, and resilience to changing rules.
\item We developed a open-source (code released soon), modular tool fully compatible with the Gymnasium interface, designed to augment existing RL ecosystems and provide a blueprint for creating dynamic, behavior-responsive environments.
\end{itemize}

By transforming static benchmarks into behavior-responsive environments, AbideGym enables the development and evaluation of agents that can adapt, generalize, and perform robustly under dynamic and unpredictable conditions.

\section{Related Work}
The MiniGrid environment \cite{minigrid} has become a widely used evaluation testbed in reinforcement learning (RL) research due to its lightweight and minimalistic design, which enables rapid prototyping and emphasizes core challenges such as sparse rewards, exploration and hierarchical planning. Despite its utility, MiniGrid and similar static benchmarks are limited in their ability to study adaptive behaviors: once an episode begins, both the environment layout and task rules remain fixed. To address these limitations, procedural content generation (PCG) has emerged as a prominent paradigm \cite{justesen2019illuminating}, with benchmarks such as CoinRun and Procgen \cite{cobbe2019} that leverage PCG to create vast, diverse training and testing levels. By exposing agents to a wide distribution of environments, these approaches encourage the development of robust policies that generalize across varied initial conditions. More recent advances have further enriched PCG through reinforcement learning and generative AI techniques, including adversarial RL-based generation \cite{rosen2022adversarial}, large language model-instructed procedural generation \cite{sun2023llm_pcg}, and multi-objective instruction-aware representation learning \cite{liu2025pcgrl}. While these methods improve robustness across episodes, the environment itself remains static within each episode, leaving intra-episode adaptation largely untested. AbideGym addresses this gap by introducing agent-aware, intra-episode dynamics, requiring online policy adaptation rather than relying solely on pre-compiled robust strategies.

Complementing this, curriculum learning (CL) provides a structured approach to improving learning efficiency and generalization by sequencing tasks from simple to complex \cite{bengio2009}. In the context of RL, CL has been operationalized through mechanisms for task generation, sequencing, and knowledge transfer \cite{narvekar2020framework}, with automatic curriculum learning (ACL) further adapting task difficulty based on agent performance to create a feedback loop between learner and environment \cite{portelas2020automatic}. Recent contributions include adaptive curriculum finetuning (AdaRFT) \cite{liu2025adarft}, Self-Evolving Curriculum (SEC) \cite{shinn2023sec}, portable curricula for visual domains \cite{lyu2024}, and curriculum-based quadrotor control \cite{sun2023quadrotor}. \emph{AbideGym extends this paradigm with an inverse curriculum, where the complexity of the environment increases in response to agent inactivity through grid resizing and additional obstacles.} The key intuition behind this is that we use agent's confidence or hesitation as a policy parameter when it comes to policy optimization. Also unlike standard curriculum learning, which typically scales difficulty upward as agents succeed, this approach actively challenges non-adaptive policies and promotes generalization across a distribution of task scales.

Real-world RL applications further necessitate agents capable of continual adaptation to evolving transition dynamics, reward structures, or task objectives \cite{padakandla2020reinforcement}. Continual learning frameworks have emerged to mitigate catastrophic forgetting \cite{kirkpatrick2017overcoming} and enable retention of previously acquired knowledge while learning new tasks. Benchmarks and architectures such as Continual World \cite{continualworld2021}, life-long visual RL with world models \cite{sun2023lifelong}, diffusion-based stable continual RL \cite{zhang2024stable}, offline video-game RL scenarios \cite{liu2025cont_offline}, and autoencoder-driven modular continual RL \cite{shinn2023cont_modular} highlight both the challenge and necessity of evaluating agents under dynamic conditions. AbideGym provides a structured, agent-aware platform to systematically test intra-episode adaptation, strategy switching, and resilience to rule changes. This allows for controlled evaluation of continual learning algorithms in a reproducible manner.

Unlike existing dynamic benchmarks that often introduce environmental changes randomly or according to pre-defined schedules, AbideGym is uniquely agent-aware: perturbations are discrete events triggered by measurable agent behaviors, such as prolonged inactivity, and their effects on the environment are well-defined and reproducible. This design allows for rigorous study of online adaptation and robust policy learning, making AbideGym a compelling tool for to evaluate models on hallucination, catastrophic forgetting and most importantly, to advance continual learning in reinforcement learning agents.

\section{AbideGym's Environment Design}
AbideGym's first environment is implemented as a modular wrapper around the MiniGrid's \cite{minigrid} DoorKey environment, introducing dynamic mechanisms that respond to an agent's behavior. It transforms a static task into a non-stationary challenge that requires policy adaptation.

\subsection{Grid Setup}
The initial environment is a grid containing \emph{walls, a key, a locked door, and a goal}. The trigger tile is placed to avoid overlap with existing objects. Upon resizing, the grid grows by 2 units per side for example (customizable parameter), and the agent is reset to the start position.

AbideGym exposes structured state information through the \texttt{EnvState} dataclass:

\begin{equation} \texttt{EnvState} = \Big( \begin{aligned} & \texttt{agent\_pos},\ \texttt{agent\_dir},\ \texttt{door\_locked},\ \texttt{has\_key}, \\ & \texttt{trigger\_color},\ \texttt{perturbation},\ \texttt{time\_step} \end{aligned} \Big) \end{equation} \textit{where:} 

\begin{align*} \texttt{agent\_pos} & = (x, y) \text{ coordinates of the agent} \\ \texttt{agent\_dir} & = \text{agent's facing direction} \\ \texttt{door\_locked} & = \text{boolean indicating door status} \\ \texttt{has\_key} & = \text{boolean if agent holds a key} \\ \texttt{trigger\_color} & = \text{color of trigger tile (if present)} \\ \texttt{perturbation} & = 1.0 \text{ if perturbation active, 0 otherwise} \\ \texttt{time\_step} & = \text{step count} \end{align*}

\subsection{Core Task and Dynamic Mechanisms}
The base environment is a classic grid-world task where an agent must pick up a key, use it to unlock a door, and reach a goal tile. AbideGym augments this with two agent-aware mechanisms:
\begin{enumerate}
    \item \textbf{Timeout-Based Perturbations:} The environment tracks the number of steps an agent has taken since its last move. If this count exceeds a configurable timeout threshold, the environment's rules are perturbed. The standard causal pathway (key unlocks door) is disabled, and a new solution path is introduced in the form of a \emph{trigger tile}. Stepping on this tile now unlocks the door.

\begin{figure}[h]
  \centering
  \includegraphics[width=\textwidth]{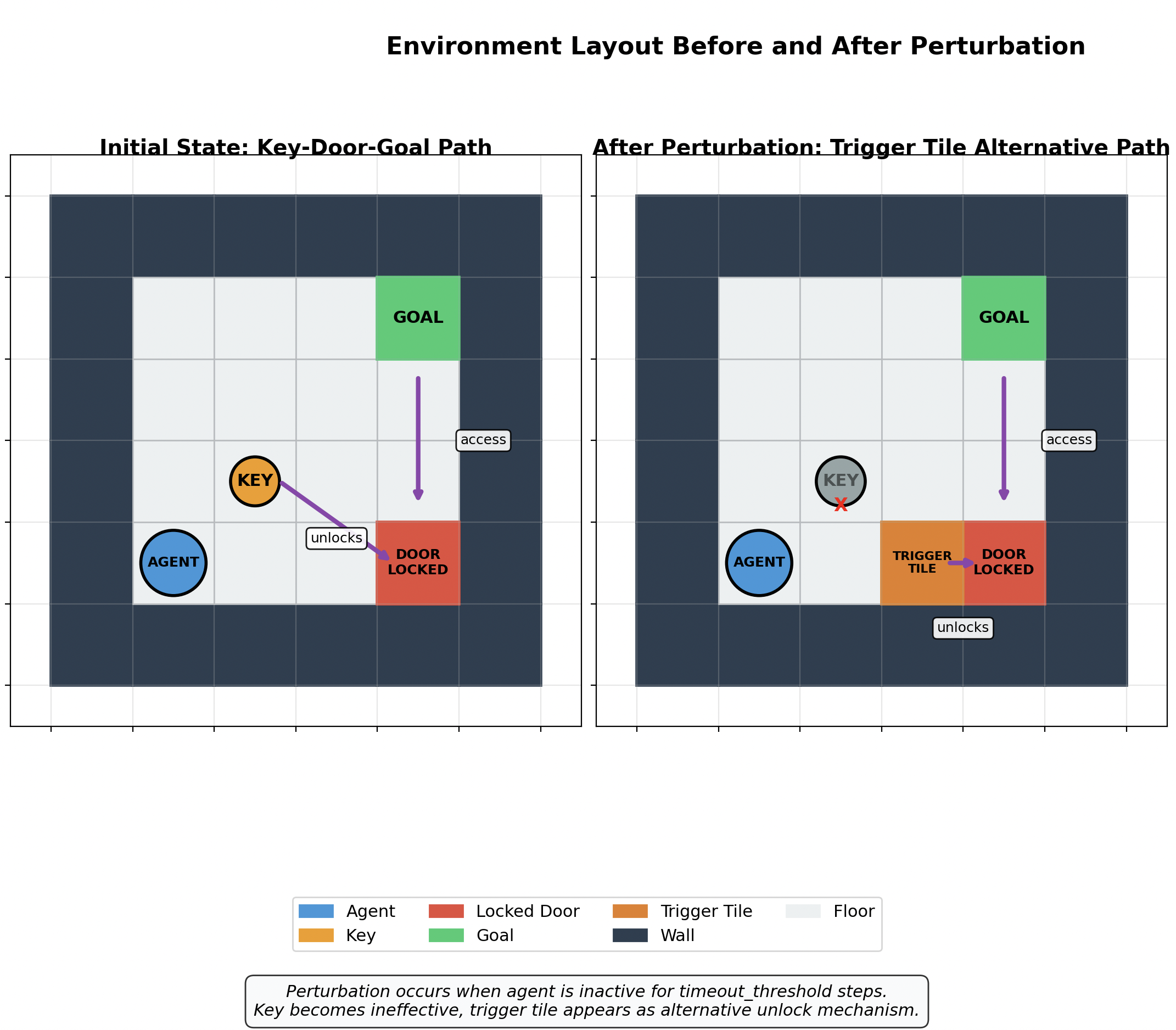}
  \caption{The AbideGym environment layout before and after a timeout-based perturbation. Left: The initial state where the agent must use the key to unlock the door. Right: After the agent remains inactive, the key becomes ineffective, and an orange trigger tile appears, offering an alternative path to the goal.}
  \label{fig:perturbation_layout}
\end{figure}

    \item \textbf{Dynamic Resizing:} If the agent remains inactive for an even longer duration, a second threshold is crossed, triggering an environment resize. The grid expands (e.g., from $4 \times 4$ to $6 \times 6$), new walls may be added, and the agent's position is reset. This escalates navigational complexity, testing the agent's ability to generalize its strategy to larger, more complex state spaces.
\end{enumerate}

\subsection{Perturbation and Adaptation Flow}
The environment's dynamics create a multi-stage challenge that encourages strategic adaptation. The typical flow of an interaction is as follows:

\paragraph{Initial State \& Key Strategy} The agent begins in a standard MiniGrid setup. The optimal policy involves navigating to the key, picking it up, moving to the door, unlocking it, and proceeding to the goal. Figure \ref{fig:perturbation_layout} (left) illustrates this initial state.

\paragraph{Warning and Perturbation} If the agent struggles and remains inactive (e.g., stuck in a loop), the environment initiates a perturbation. When \texttt{steps\_since\_move} exceeds half of the \texttt{timeout\_threshold}, a warning phase begins. If the threshold is fully exceeded, the perturbation is triggered: the key no longer functions, and the trigger tile appears. Figure \ref{fig:perturbation_layout} (right) shows the environment after this change, presenting the agent with a new problem to solve.

\begin{figure}[h]
  \centering
  \includegraphics[width=\textwidth]{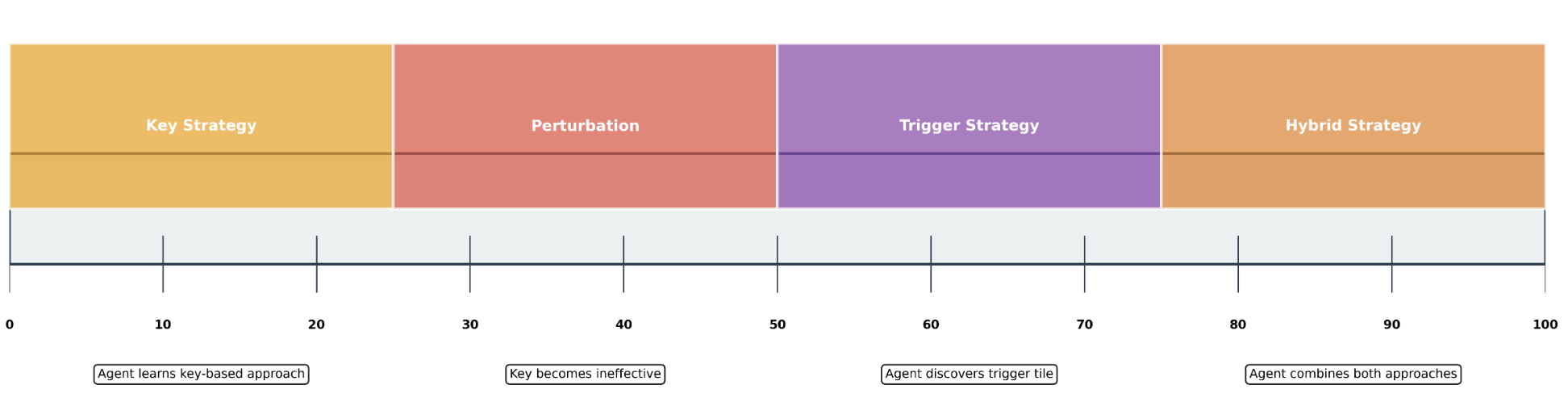}
  \caption{An idealized timeline of an agent's strategy evolution in AbideGym. The agent progresses from learning a single strategy to discovering an alternative after a perturbation, and eventually to a hybrid approach that flexibly combines both.}
  \label{fig:strategy_evolution}
\end{figure}

\paragraph{Strategy Evolution} Over multiple episodes, an adaptive agent is expected to evolve its strategy. Figure \ref{fig:strategy_evolution} depicts this ideal learning trajectory. The agent first masters the standard \emph{Key Strategy}. After encountering perturbations, it learns the \emph{Trigger Strategy}. Ultimately, a sophisticated agent might develop a \emph{hybrid strategy} efficiently choosing between the key or the trigger based on their relative positions and the current state of the environment.

\begin{figure}[h]
  \centering
  \includegraphics[width=\textwidth]{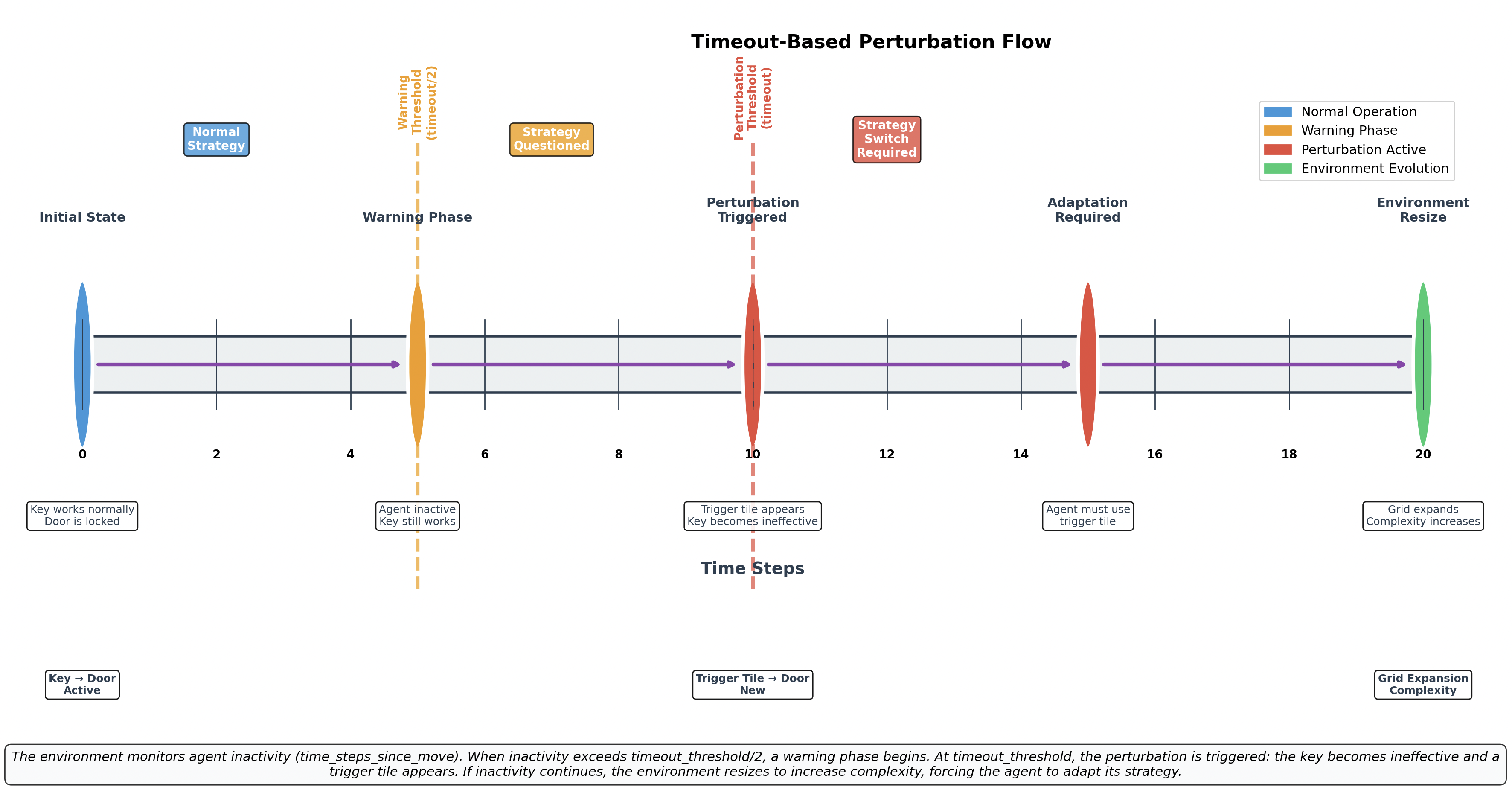}
  \caption{A detailed timeline of the timeout-based perturbation flow. The environment monitors agent inactivity, enters a warning phase, and triggers a perturbation that changes the task rules, forcing the agent to adapt its strategy.}
  \label{fig:perturbation_flow}
\end{figure}

Figure \ref{fig:perturbation_flow} provides a detailed step-by-step timeline of a single perturbation event, illustrating the transitions from normal operation to the final adaptation requirement.

\subsection{Dynamic Resizing for Generalization}
\begin{figure}[h]
  \centering
  \includegraphics[width=\textwidth]{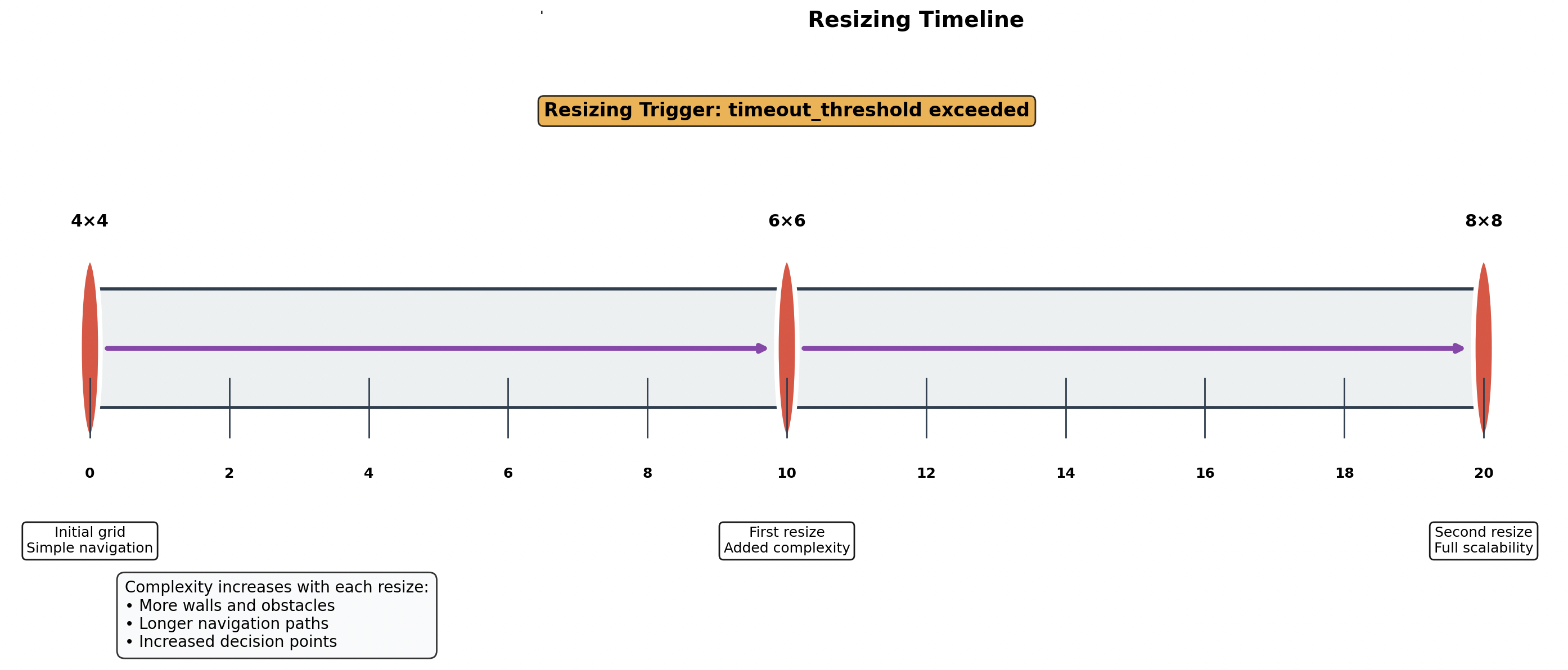}
  \caption{The resizing timeline, illustrating how prolonged agent inactivity triggers sequential increases in the grid size, escalating the navigational complexity of the task.}
  \label{fig:resizing_timeline}
\end{figure}

If an agent fails to adapt to a perturbation and remains inactive, the dynamic resizing mechanism is triggered. This feature is designed specifically to test and promote policy generalization and functions as an \emph{inverse curriculum}. As shown in the timeline in Figure \ref{fig:resizing_timeline}, repeated failures lead to a more complex and progressively larger environment.

\begin{figure}[h]
  \centering
  \includegraphics[width=\textwidth]{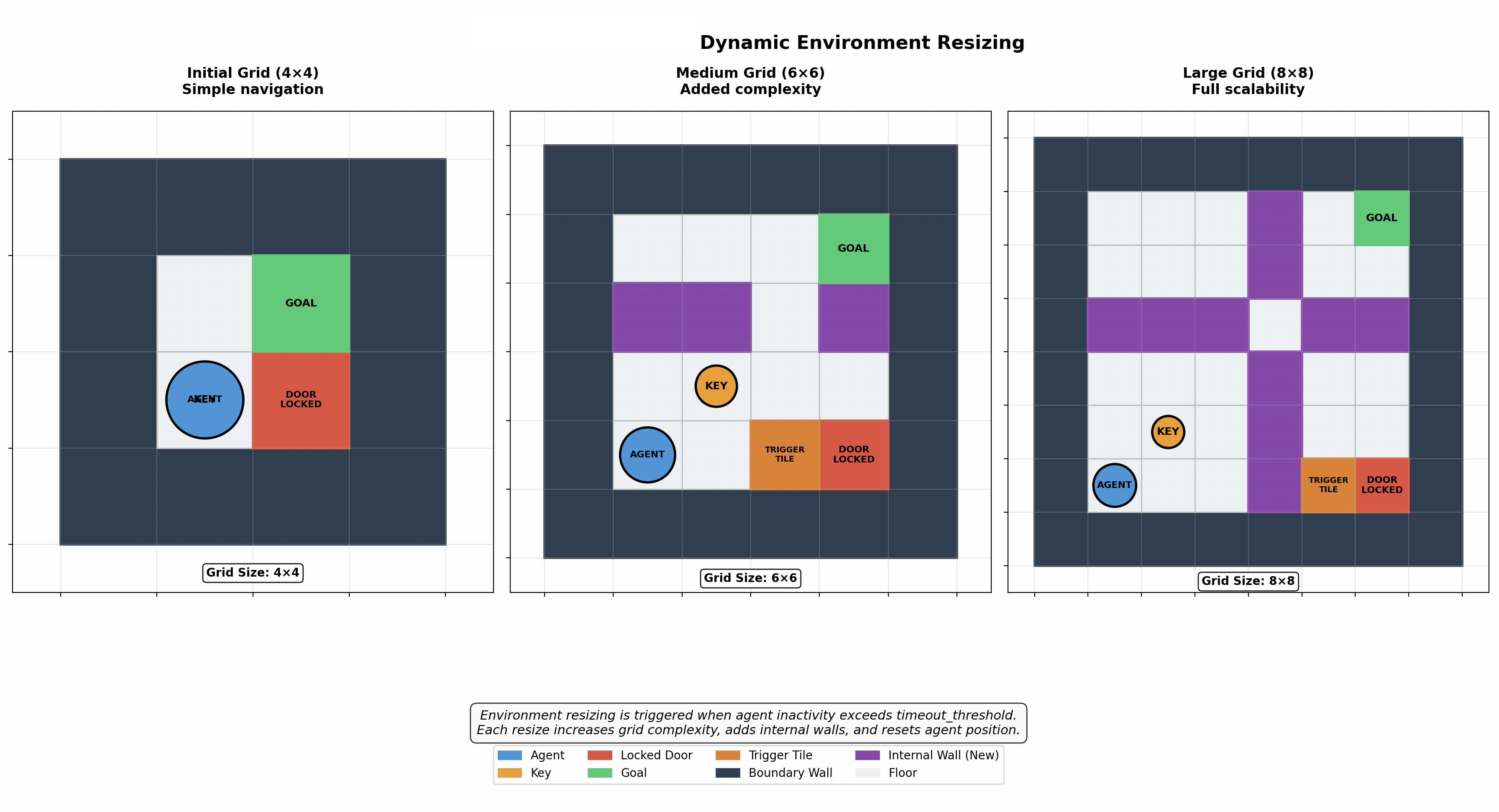}
  \caption{Visual representation of dynamic environment resizing. The grid expands from a simple $4 \times 4$ layout to more complex $6 \times 6$ and $8 \times 8$ configurations with internal walls, challenging the agent's ability to generalize its navigation policy.}
  \label{fig:resizing_visualization}
\end{figure}

Figure \ref{fig:resizing_visualization} provides a visual comparison of the grid at different stages of resizing. The increase in size is accompanied by the addition of internal walls, creating more complex environments that require more sophisticated exploration and planning policies. By making the environment harder in response to failure, AbideGym discourages agents from memorizing simple paths and instead pushes them to learn more generalizable navigation strategies.

\subsection{State Representation and Implementation}
AbideGym exposes a structured state representation that provides clear, interpretable information about the environment's status. Table \ref{tab:state_variables} details the key state variables. These variables are categorized into three groups, categorized by their role in the dynamic mechanisms. \emph{Perturbation Variables} control the state of the task rules and dynamic changes to the task rules. \emph{Adaptation Variables} track agent behavior and trigger the dynamic changes like resizing and Neutral variables provide basic state information.

\begin{table}[h]
  \centering
  \caption{State variables exposed by AbideGym}
  \label{tab:state_variables}
  \begin{tabular}{llll}
    \toprule
    \textbf{State Variable} & \textbf{Role} & \textbf{Description} & \textbf{Impact on Environment} \\
    \midrule
    \texttt{agent\_pos}        & Perturbation & Agent coordinates       & Determines spatial context \\
    \texttt{agent\_dir}        & Neutral      & Facing direction        & Navigation orientation \\
    \texttt{door\_locked}      & Perturbation & Door lock status        & Blocks goal access \\
    \texttt{has\_key}          & Perturbation & Key possession          & Controls door unlocking \\
    \texttt{trigger\_color}    & Perturbation & Trigger tile color      & Indicates perturbation \\
    \texttt{perturbed}         & Perturbation & Perturbation active     & Environment state flag \\
    \texttt{time\_step}        & Adaptation   & Current time            & Temporal progression \\
    \texttt{steps\_since\_move}& Adaptation   & Inactivity counter      & Resizing trigger \\
    \texttt{current\_size}     & Adaptation   & Grid dimensions         & Environment scale \\
    \texttt{timeout\_threshold}& Adaptation   & Timeout limit           & Resizing threshold \\
    \bottomrule
  \end{tabular}
\end{table}

One of the core reasons for developing AbideGym as a wrapper around existing environments was to ensure seamless integration with existing RL libraries and research pipelines. Although since this is still early work, our current implementation only covers Gymnasium \cite{gymnasium} API only. Table \ref{tab:env_comparison} summarizes how we envision AbideGym's features to create a distinct evaluation benchmark compared to the existing benchmarks.

\begin{table}[h]
  \centering
  \caption{Comparative analysis of RL generalization environments.}
  \label{tab:env_comparison}
  \begin{tabular}{@{}p{3.8cm} p{2cm} p{3cm} p{3.4cm}@{}}
    \toprule
    \textbf{Feature} & \textbf{MiniGrid (Baseline)} & \textbf{PCGRL} & \textbf{AbideGym} \\
    \midrule
    Environment Dynamics & Static & Static within episode & Dynamic within episode \\
    Source of Variation  & None & Procedural Generation & Agent Inactivity Triggers \\
    Change Timing        & N/A  & Inter-Episode & Intra-Episode \\
    Type of Change       & N/A  & Layout, Visuals & Task Rules, Layout, Complexity \\
    Primary Challenge    & Exploration, Planning & Policy Generalization & Policy Adaptation \& Generalization \\
    \bottomrule
  \end{tabular}
\end{table}
\section{Discussion}
The design of AbideGym is guided by several core principles that distinguish it from conventional RL benchmarks for agentic systems.

\paragraph{Adaptability over Simplification} A key design choice in AbideGym is to increase environmental complexity when an agent is struggling. Although this may seem counterintuitive at first, as typical curriculum learning frameworks aim to simplify tasks to facilitate learning. However, the objective of AbideGym is not to help an agent solve one specific task more easily. Instead, its goal is to test and cultivate the agent's \emph{adaptability}. The core intuition behind this was mismatched states in multi-agent systems or syncing issues. By altering the rules and expanding the state space in response to failure as well as hesitation, the environment actively discourages brittle, memorized strategies. It forces the agent to develop a more general and robust policy that can cope with variability and unexpected change, mirroring the demands of real-world agentic systems.

\paragraph{A General Framework for Dynamic Environments} While this paper presents an implementation based on MiniGrid, the underlying architecture of AbideGym is intended as a general framework. The core components: \emph{behavioral monitoring} (e.g., inactivity timers), \emph{event-triggered perturbations}, and {dynamic parameter adjustment} (e.g., resizing), can be implemented as a wrapper around any standard RL environment. The goal is not to replace existing libraries like Gymnasium or other even text and reward-based environments, but to expand their capabilities, allowing researchers to easily inject controlled non-stationarity and perturbation into a wide range of tasks. This is driven by the key motivation that most of the benchmarks get reverse engineered very quickly, thus making true evaluation incredibly hard and SFT will remain a core challenge because effectively labelling data continues to remain a open challenge, and especially more so for agentic systems in production.

\paragraph{Perturbations as Causal Breaks} The mechanisms in AbideGym can be viewed through the lens of causal reasoning. An agent learning the standard task develops an implicit causal model: picking up the key \emph{causes} the agent to possess the key, which in turn \emph{enables} the action of unlocking the door. The timeout-based perturbation is a direct intervention on this causal graph; it severs the link between possessing the key and being able to unlock the door. This event creates a \emph{causal break}, a mismatch between the agent's learned model of the world and the observed reality. To succeed, the agent must perform a form of causal discovery: it must infer that its old model is invalid and identify the new causal pathway to its goal (i.e., stepping on the trigger tile now causes the door to unlock). Infact, we developed \emph{AbideGym environments for developing and evaluating agents capable of causal inference and reflective reasoning}, as proposed in the Causal Reflection framework\cite{aryan2025}.

\section{Future Work and Broader Impact}
AbideGym is still an early implementation, currently limited in scope to object manipulation tasks only. The core reason for this was to evaluate our upcoming work in Causal Reflection Agents and the interpretability of such environments. But extending it to text-based environments and other RL environments will be explored in future work. 

We have also identified several technical challenges that would help accelerate research in this area for the research community, some of them being:

\subsection{Going from Heuristic-Based to Competence-Based Curricula} 
The current triggers for perturbation and resizing are based on a simple heuristic for failure (inactivity). A natural extension would be to develop more sophisticated, competence-based triggers. For example, perturbations could be activated by metrics of agent uncertainty, such as low policy entropy (indicating over-exploitation) or high variance in value function estimates (indicating confusion). This would create a truly adaptive curriculum where the environment's challenges are tailored to the agent's internal cognitive state, not just its external behavior.

\subsection{Developing Model Benchmarks for Continual Learning and Catastrophic Forgetting} 
A standard experimental protocol could involve training an agent to mastery on the static key-door task, then transferring it to an AbideGym environment with frequent perturbations. The key research question would be whether the agent can learn the new trigger-tile strategy without catastrophically forgetting the original key-based strategy. This provides a controlled setting to evaluate the efficacy of continual learning algorithms \cite{kirkpatrick2017overcoming} in preventing knowledge degradation. We believe benchmarking different models for their ability to do so using AbideGym would emerge as a strong evaluation metric for the OSS community.

\paragraph{Meta-Learning for Adaptation} The parameterized nature of AbideGym (e.g., \texttt{timeout\_threshold}, resizing increments) makes it an ideal environment for meta-reinforcement learning \cite{finn2017model}. Future work can include how an agent could be trained across a distribution of AbideGym environments with different dynamic parameters. The objective would be to meta-learn an adaptive policy or learning algorithm that can quickly identify and adapt to the specific rules of a new, unseen AbideGym configuration within a few interactions; in essence, to "learn how to adapt".

\paragraph{Stochastic and Multi-Agent Extensions} Future versions of the framework could incorporate more complex dynamics. Perturbations could be stochastic rather than deterministic, occurring with a certain probability after a timeout, which would force the agent to reason under uncertainty. Furthermore, the framework could be extended to multi-agent scenarios, where the actions (or inaction) of one agent trigger environmental changes that affect all agents, creating a rich evaluation testbed for studying multi-agent coordination, competition, and adaptation. 

By continuing to build upon AbideGym in any of these directions, we can develop increasingly sophisticated benchmarks that push the boundaries of agent adaptability and generalization.

\section{Conclusion}
We have introduced AbideGym, a dynamic environment framework designed to address a critical gap in reinforcement learning research: the need for benchmarks that test an agent's ability to adapt to intra-episode changes. By integrating timeout-based perturbations and dynamic resizing into the classic MiniGrid environment, AbideGym creates a controlled, non-stationary setting that challenges agents to revise failing strategies and generalize across varying levels of complexity. By facilitating research in continual learning, meta-learning, and causal reasoning, AbideGym offers a practical path toward creating AI systems that can learn not only what to do, but how to adapt when the rules of the world change.

\medskip
\bibliographystyle{unsrtnat}
\bibliography{references}

\end{document}